\documentclass[letterpaper, 10 pt, conference]{ieeeconf}  

\IEEEoverridecommandlockouts                              

\usepackage{cite}
\usepackage{amsmath,amssymb,amsfonts}
\usepackage{algorithm}
\usepackage{algpseudocode}
\usepackage{graphicx}
\usepackage{textcomp}
\usepackage{xcolor}
\usepackage{array}
\usepackage{booktabs}
\newcolumntype{C}[1]{>{\centering\arraybackslash}m{#1}}

\title{\LARGE \bf
Safety-aware Skill Adaptation for Reinforcement Learning in Dynamic Environments
}

\author{A K M Nadimul Haque, Sheila Sutjipto, Marc G. Carmichael and Teresa Vidal-Calleja 
\thanks{This work was supported by the Industrial Transformation Training Centre (ITTC) for Collaborative Robotics in Advanced Manufacturing (also known as the Australian Cobotics Centre) funded by ARC (Project ID: IC200100001).}
\thanks{All authors are with the Robotics Institute, University of Technology Sydney, Australia (e-mail: akmnadimul.haque@student.uts.edu.au, \{sheila.sutjipto,marc.carmichael,teresa.vidalcalleja\}@uts.edu.au)}
}

\begin{document}

\maketitle
\thispagestyle{empty}
\pagestyle{empty}


\begin{abstract}

Skill adaptation frameworks based on reinforcement learning often require restrictive assumptions to maintain stability, such as fixed observations or tightly controlled exploration schedules. In cluttered and dynamic environments, however, unrestricted exploration can lead to unsafe behaviour and unstable learning, particularly when task-relevant observations lie near obstacles or involve moving objects. In this work, we present Dist-GPRL, a distance-aware and safety-guided reinforcement learning framework for structured robot skill adaptation. Building upon Gaussian Process (GP)-based skill parameterisation, our framework sequentially adapts overlapping local windows of sparse trajectory via-points rather than modifying the complete skill at every policy step. Raw policy outputs are correlated through the GP covariance structure, producing temporally coherent trajectory updates while reducing the action-space and credit-assignment difficulties associated with global trajectory adaptation. Safety is incorporated through two complementary forms of guidance. A safe-subspace prior derived from the Hausdorff Approximation Planner (HAP) biases policy exploration toward feasible regions, while dynamically updated distance field clearance and gradient rewards provide local obstacle awareness. A trajectory-kinematics similarity regulariser further preserves the demonstrated velocity and acceleration characteristics during adaptation. We evaluate the framework on two dynamic object-manipulation tasks in simulation and transfer the learned policy to real-world robot execution. Experimental results demonstrate higher task success, lower collision frequency, and more stable learning than the baselines, while preserving the kinematic characteristics of the demonstrated skill.

\end{abstract}

\begin{figure}
    \centering
    \includegraphics[width=0.9\linewidth]{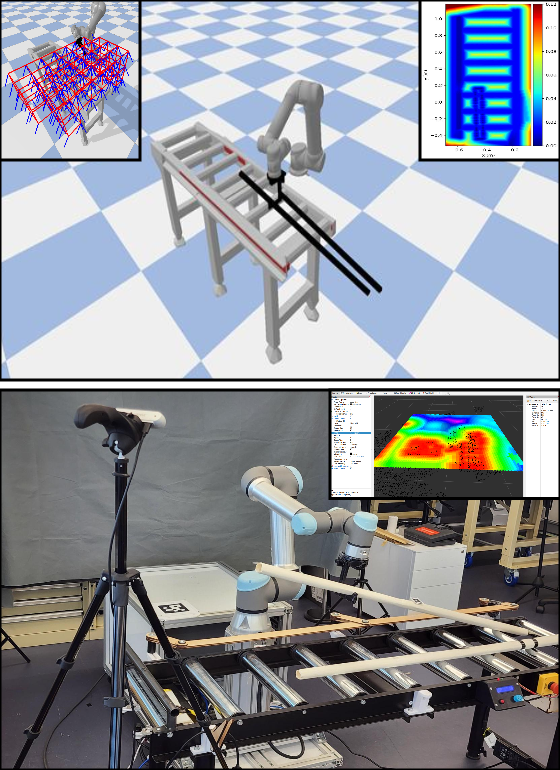}
    \caption{Distance field guidance with safe subspace prior facilitates policies that produce structured trajectory modifications that avoid obstacles while preserving demonstrated kinematics. The learned policy adapts skills in simulation (top) and transfers them to real-world robot execution (bottom).}
    \label{fig:teaser}
\end{figure}
\section{Introduction}
Collaborative robots operating in industrial and unstructured environments must adapt learned skills to varying task configurations (TC) such as different object poses, and start or end conditions, while maintaining safety and motion consistency. Rather than simply reproducing demonstrated trajectories, modern robotic systems increasingly focus on learning parameterised skills that can be robustly adapted to new TCs while retaining the kinematic properties of expert demonstrations. Skill learning encodes structured motion priors, whereas safe skill adaptation requires generalising these priors to novel and potentially dynamic task configurations without compromising safety or motion consistency.

Recent works have explored hybrid frameworks that combine probabilistic skill parameterisation with deep reinforcement learning (DRL) to enable efficient adaptation. Representations such as Dynamic Movement Primitives (DMPs)~\cite{davchev2022residual}, and Gaussian Mixture Models (GMMs)~\cite{nematollahi2022robot} have been integrated with DRL. However, DMPs lack the ability to model trajectory distributions, while GMM-based approaches often exhibit limited expressiveness in complex manipulation tasks. ProMPs~\cite{paraschos2013probabilistic} are more expressive, but lack intuitive integration with DRL adaptation. Gaussian Processes (GPs) also provide a more expressive alternative for skill representation and have been integrated with DRL frameworks for adaptation~\cite{haque2026robotskilllearningadaptation}, enabling structured adaptation on TCs and improved retention of demonstrated kinematics. Despite these advantages, existing GP-based adaptation frameworks rely on restrictive assumptions to maintain stability, such as fixed and collision-free TC observations, as well as annealed exploration. These assumptions break down in cluttered environments, where TC observations may lie near obstacles and unrestricted policy exploration can lead to unstable learning.

Specifically, ensuring safety during structured skill adaptation remains largely underexplored. While some approaches incorporate safety through reward shaping, explicit geometry-aware exploration remains limited. Prior work has introduced task-dependent safe subspaces~\cite{haque2024constrained} to guide policy exploration toward feasible robot configurations. However, constructing and updating these subspaces can be computationally expensive and robot-specific, limiting their applicability when the environment changes dynamically.

A complementary notion of safety can be expressed through proximity to obstacles. Distance fields provide a robot-agnostic and continuous representation of clearance from nearby surfaces, together with well-defined gradients that indicate locally safer directions for motion. They have proven useful in safe navigation~\cite{long2021learning} and manipulation~\cite{liu2023safe}. However, their integration with structured adaptation of demonstrated skills remains relatively unexplored.

In this work, we propose Dist-GPRL, a safety-aware reinforcement learning framework that extends structured GP-based skill adaptation to cluttered and dynamic environments. Unlike prior GPRL formulations~\cite{haque2026robotskilllearningadaptation}, which adapt the complete via-point set simultaneously, Dist-GPRL sequentially updates overlapping local trajectory windows. This reduces the effective action space and improves credit assignment while retaining the global GP trajectory representation. Within each window, raw policy outputs are transformed through the GP covariance rather than applied independently to individual via-points. This couples neighbouring modifications according to the demonstrated trajectory prior, producing temporally coherent adaptations and improving learning stability. Safety is introduced through two complementary forms of guidance: a safe-subspace prior derived from the Hausdorff Approximation Planner (HAP) biases policy exploration toward feasible pose regions, while dynamically updated distance field rewards provide local clearance and obstacle-avoidance feedback as the environment changes. Cosine similarity regularisation further preserves the demonstrated kinematic profile, preventing safety-driven adaptation from unnecessarily altering the underlying motion characteristics. Fig.~\ref{fig:teaser} illustrates Dist-GPRL applied to the dynamic bar manipulation task in both simulation and real-world execution.

To summarise, our contributions are:
\begin{itemize}

\item A local, structured skill adaptation formulation that sequentially updates overlapping temporal windows, reducing the action-space and credit-assignment difficulties associated with global via-point adaptation.

\item A GP covariance-based action transformation that maps independent policy outputs into coherent via-point modifications consistent with the temporal structure of the demonstrated skill.

\item A complementary safety-guidance formulation that combines a HAP-derived safe-subspace prior for efficient exploration, and dynamically updated distance field rewards for local obstacle awareness, while retaining demonstrated kinematics through kinematic similarity regularisation.

\item Experimental validation on dynamic manipulation tasks in simulation and on real hardware with ablation studies and safety metrics.

\end{itemize}







\begin{figure*}
    \centering
    \includegraphics[width=1\linewidth]{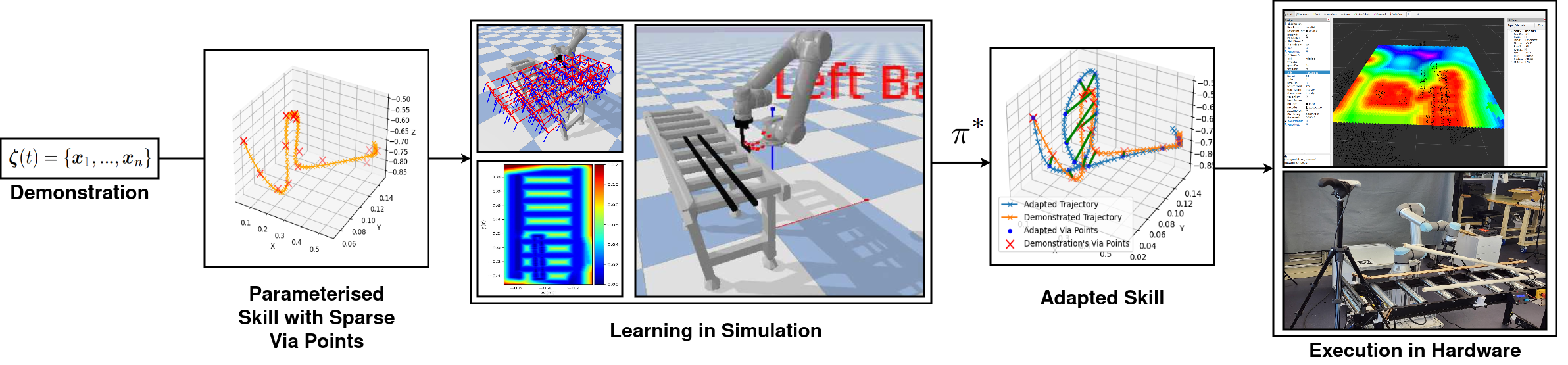}
    \caption{A demonstrated skill is parameterised with GPs and sparse via-points. During simulation training, Dist-GPRL sequentially adapts local via-point windows using covariance-transformed actions, HAP safe-subspace regularisation, and distance field safety feedback. The learned policy is then transferred to hardware to perform one-shot skill adaptation using the real-world distance field.}
    \label{fig:pipeline}
\end{figure*}
\section{Related Work}
Skill learning and consequent adaptation have gained significant traction in recent times. 
Several trajectory parameterisation frameworks have been proposed for skill representation, including Dynamic Movement Primitives (DMPs)~\cite{ijspeert2013dynamical}, Probabilistic Movement Primitives (ProMPs)~\cite{paraschos2013probabilistic}, and Gaussian Mixture Models (GMMs)~\cite{reynolds2009gaussian}. Among these approaches, Gaussian Processes (GPs)~\cite{barfoot2014batch} provide a particularly expressive representation and are well suited for structured adaptation due to their analytical derivatives and probabilistic formulation~\cite{haque2026robotskilllearningadaptation}.
The work of \cite{haque2026robotskilllearningadaptation} used a via-point-based GP parameterisation to successfully adapt the demonstrated skill to cater for a wide range of TCs. To keep the RL-based adaptation structured, the authors proposed signature similarity-based rewards and introduced heavy exploration annealing. The framework also keeps the via-point corresponding to the observed TC fixed during optimisation to aid convergence.
However, anchoring the adaptation around the fixed via-point leads to exploration only near the conditioned mean. Thus, if the observed TC is close to collision, limited exploration will result in adaptation policies that result in collisions.

Existing literature has also prioritised safety in skill learning and adaptation. Authors in \cite{haque2024constrained} used HAP \cite{sukkar2025multi} in conjunction with a GMM parameterised skill and consequent RL adaptation to encourage exploration in a safe subspace. HAP provides a subspace of poses that are safe, feasible and optimal for the robot under the current task configuration, which can effectively guide policies to converge quickly. However, building and updating dense HAP subspaces is computationally expensive. Thus, under dynamic scenarios, such notions of safe exploration become less reliable. Safety during manipulation can also be represented through geometry-aware measures of obstacle proximity. Distance fields provide such a representation by encoding the clearance between the robot and surrounding obstacles across the workspace. They are computationally more efficient, and with recent works focusing on real-time generation and updates \cite{ali2024interactive}, they provide a viable framework for manipulation in dynamic environments. 

Despite this, distance fields in combination with DRLs for skill learning have rarely appeared in the literature. Authors in \cite{zhang2025reactive} proposed a hierarchical approach where RL proposed a high-level, velocity-based planning, while a lower-level quadratic program constrained by a signed distance field (SDF) ran strict joint velocity controls. Liu et al. \cite{liu2023safe} extended the ATACOM framework \cite{liu2022robot} with a regularised deep SDF and soft actor-critic (SAC) network. The SAC policy is trained on a constrained tangential space. Strict collision constraints modelled by the ReSDF were projected to the null space for constrained policy learning. However, structured skill adaptation with distance field–based shaping within the reinforcement learning reward has been less explored, especially when combined with probabilistic trajectory representations such as GPs that enable dense action correlation and kinematic retention.

\section{Background}

\subsection{Gaussian Process Skill Parameterisation}

Gaussian Processes (GPs) provide a non-parametric probabilistic representation of continuous functions. 
A demonstrated trajectory $\boldsymbol{\zeta}(t)$ can be modelled as
\begin{equation}
\boldsymbol{\zeta}(t) \sim \mathcal{GP}(0, k(t,t')),
\end{equation}
where $k(t,t')$ is a kernel encoding temporal correlations. Given noisy observations $\mathbf{y}_i = \boldsymbol{\zeta}(t_i) + \eta$ with $\eta \sim \mathcal{N}(0,\sigma_y^2)$, the posterior mean at a query time $\hat t$ is
\begin{equation}
\boldsymbol{\zeta}(\hat t)
=
\mathbf{k}_t(\hat t)
\left[
\mathbf{K}_{tt} + \sigma_y^2 \mathbf{I}
\right]^{-1}
\mathbf{y},
\end{equation}
where $\mathbf{k}_t(\hat t)$ denotes the covariance vector between the query time $\hat t$ and the observation times $t_i$, $\mathbf{K}_{tt}$ is the covariance matrix evaluated at the observation times.

A sparse representation can be obtained by conditioning the GP on a reduced set of via-points $\boldsymbol{\Gamma}$, which act as control variables for the trajectory. 
Updates to these conditioning variables induce globally smooth trajectory modifications due to the kernel structure. 
Furthermore, derivatives of the GP posterior can be computed analytically \cite{sarkka2011linear}, providing direct access to velocity and acceleration profiles.

\subsection{Soft Actor-Critic}
Soft Actor-Critic (SAC)~\cite{haarnoja2018soft} is a stochastic, off-policy reinforcement learning algorithm designed for continuous action spaces. SAC learns a policy by maximising the expected return while incorporating an entropy regularisation term to encourage exploration. The actor's objective takes the form
\begin{equation}
\mathcal{L}_\pi
=
\mathbb{E}
\left[
\alpha \log \pi(\mathbf{a}|\mathbf{s})
-
Q(\mathbf{s},\mathbf{a})
\right],
\end{equation}
where $\mathbf{s}$ denotes the state, $\mathbf{a}$ the action sampled from the policy $\pi$, $Q(\mathbf{s}, \mathbf{a})$ is the state–action value function, and $\alpha$ is the temperature parameter controlling the strength of entropy regularisation. Entropy regularisation improves stability and exploration in high-dimensional continuous control problems.
\subsection{Euclidean Distance Fields}
A Euclidean Distance Field (EDF) is a scalar field that models the surrounding environment as a surface or manifold, $\mathcal{O} \subset \mathbb{R}^3$. Its boundary given by obstacles in the environment is defined by $\delta \mathcal{O}$ with an orientation dictated by its surface normals. The distance is given by,
\begin{equation}
d(\mathbf{x})
=
\min_{\mathbf{o} \in \delta\mathcal{O}}
\|\mathbf{x}-\mathbf{o}\|_2,
\end{equation}
where $\|\cdot\|_2$ denotes the Euclidean norm. The scalar field $d(\mathbf{x})$ is non-negative in free space and approaches zero at contact. Its gradient, $\nabla d(\mathbf{x}) =
\frac{\partial d(\mathbf{x})}{\partial \mathbf{x}}$,  points in the direction of maximal increase of clearance and thus provides a locally optimal obstacle-avoidance direction. Due to their smoothness and differentiability properties, distance fields are widely used in motion planning and optimisation-based control.

\section{Dist-GPRL: Safety-aware Skill Adaptation}
This section presents the formulation for the proposed safety-aware skill adaptation pipeline. An overview of our approach is shown in Fig.~\ref{fig:pipeline}.

\subsection{Problem Formulation}

We first formulate the skill adaptation problem similar to the work of~\cite{haque2026robotskilllearningadaptation}. Let us consider a demonstrated skill 
$\boldsymbol{\zeta}(t) = \{(\mathbf{p}_1,\mathbf{r}_1), (\mathbf{p}_2, \mathbf{r}_2), \ldots, (\mathbf{p}_n, \mathbf{r}_n)\}$, 
where $\mathbf{p}_i \in \mathbb{R}^3$ and $\mathbf{r}_i \in \mathbb{R}^3$, parameterised with six separate GPs using a sparse set of via-points as
\begin{equation}
    \begin{aligned}
    \mathbf{p}(t) &= \mathbf{k}_{t}(t) \left[\mathbf{K}_{tt} + \sigma_y^2 \mathbf{I} \right]^{-1} \boldsymbol{\Gamma}_p, \\
    \mathbf{r}(t) &= \mathbf{k}_{t}(t) \left[\mathbf{K}_{tt} + \sigma_y^2 \mathbf{I} \right]^{-1} \boldsymbol{\Gamma}_r,
    \end{aligned}
\label{eq:gp_param}
\end{equation}
where $\boldsymbol{\Gamma}_p$ and $\boldsymbol{\Gamma}_r$ are the via-point vectors for position and orientation, respectively. Let $\boldsymbol{\Gamma}_{t_s} \in \mathbb{R}^{N \times 6}$ denote the full via-point set at RL step $t_s$, where $N$ denotes the total number of via-points. Given a new TC, the agent must find an adapted trajectory $\boldsymbol{\zeta}'(t)$ through modifications of these via-points $\Delta\boldsymbol{\Gamma}$ such that:
\begin{enumerate}
    \item \textbf{Kinematics retention:} the demonstrated velocity profile is preserved,
    \(
    \min \| \dot{\boldsymbol{\zeta}}'(t) - \dot{\boldsymbol{\zeta}}(t) \|.
    \)
    \item \textbf{Safety:} the adapted trajectory remains collision-free,
    \(
    d(\mathbf{p}(t)) > 0,\ \forall t,
    \)
    where $d(\cdot)$ denotes a differentiable EDF.
\end{enumerate}
We formulate the skill adaptation as an MDP 
\(
\mathcal{M} = (\mathcal{X}, \mathcal{A}, \mathcal{P}, r, \gamma),
\)
where the agent observes and only updates a local subset of via-points at each RL step. A phase variable $\phi_{t_s} \in [0,1]$ determines an active temporal window 
$\mathcal{W}_{t_s} \subset \{1,\ldots,N\}$ with $|\mathcal{W}_{t_s}|=L$, where $L$ denotes the window size. Each RL time step $t_s$ corresponds to the adaptation of the contiguous trajectory segment defined by $\mathcal{W}_{t_s}$.

At time step $t_s$, the state $\mathbf{s}_{t_s} \in \mathcal{X}$ is defined as
\begin{equation}
\mathbf{s}_{t_s} =
\left[
\boldsymbol{\Gamma}_{t_s}^{\mathcal{W}_{t_s}},
\mathbf{c}_{t_s},
\mathbf{e}_o(t_s),
\mathbf{e}_g(t_s),
\phi_{t_s}
\right],
\label{eq:RL_state}
\end{equation}
where $\boldsymbol{\Gamma}_{t_s}^{\mathcal{W}_{t_s}} \in \mathbb{R}^{L\times6}$ are the windowed via-points, $\mathbf{c}_{t_s} \in \mathbb{R}^{L}$ are the corresponding EDF clearances evaluated at the windowed via-points, $\mathbf{e}_o(t_s)$ denotes the distance vector from end-effector to the object to be manipulated, and $\mathbf{e}_g(t_s)$ is the distance vector from end-effector to the goal.

At each time step, the policy outputs a raw corrective action $\tilde{\mathbf{a}}_{t_s} \in \mathbb{R}^{L\times6}$ for the active window. To correlate updates using the GP prior, we precompute a global correlation operator
\begin{equation}
\mathbf{S} = \mathbf{K}_{tt}\left(\mathbf{K}_{tt} + \lambda \mathbf{I}\right)^{-1}, 
\quad \text{where } \mathbf{K}_{tt}\in\mathbb{R}^{N\times N},
\label{eq:global_S}
\end{equation}
and apply the principal submatrix $\mathbf{S}_{\mathcal{W}_{t_s}} \in \mathbb{R}^{L\times L}$ indexed by $\mathcal{W}_{t_s}$ to obtain the correlated corrective actions
\begin{equation}
\Delta \boldsymbol{\Gamma}_{t_s} = \mathbf{S}_{\mathcal{W}_{t_s}}\,\tilde{\mathbf{a}}_{t_s},
\label{eq:transformed_action}
\end{equation}
where $\lambda>0$ controls the strength of action correlation. 
Thus, the via-points evolve as
\begin{equation}
\label{eq:updated_via_points}
\boldsymbol{\tilde\Gamma}_{t_s}^{\mathcal{W}_{t_s}} =
\boldsymbol{\Gamma}_{t_s}^{\mathcal{W}_{t_s}} + \Delta \boldsymbol{\Gamma}_{t_s},    
\end{equation}
where the updated via-points $\boldsymbol{\tilde\Gamma}_{t_s}^{\mathcal{W}_{t_s}}$ define the adapted trajectory $\boldsymbol{\zeta}'(t)$ through Eq.~\eqref{eq:gp_param}.

\subsection{Action Biasing with Cosine Similarity and HAP Prior}

To preserve the demonstrated kinematic structure and guide exploration toward safe regions, we augment the SAC actor objective with two regularisation terms: (i) a cosine similarity loss on trajectory kinematics, and (ii) a HAP-based safe-subspace prior.
\paragraph{Cosine Similarity Loss}
Given the adapted trajectory $\boldsymbol{\zeta}'(t)$ under via-point updates $\Delta \boldsymbol{\Gamma}_{t_s}$, we query its velocity and acceleration profiles using the GP derivative maps.
Let $\dot{\boldsymbol{\zeta}}(t), \ddot{\boldsymbol{\zeta}}(t)$ denote the demonstrated velocity and acceleration, and $\dot{\boldsymbol{\zeta}}'(t), \ddot{\boldsymbol{\zeta}}'(t)$ the adapted counterparts. We define the similarity loss over trajectory time as
\begin{equation}
\begin{aligned}
\mathcal{L}_{\mathrm{sim}}
&=
\lambda_v
\left(
1 - \frac{1}{T}\sum_{t}
\cos\!\left(\dot{\boldsymbol{\zeta}}'(t), \dot{\boldsymbol{\zeta}}(t)\right)
\right) \\
&\quad +
\lambda_a
\frac{1}{T}\sum_{t}
\left\|
\ddot{\boldsymbol{\zeta}}'(t) - \ddot{\boldsymbol{\zeta}}(t)
\right\|_1 .
\end{aligned}
\label{eq:sim_compact}
\end{equation}
where $T$ denotes the number of queried trajectory samples. This term constrains the policy to retain the demonstrated motion directionality and smoothness while adapting via-points.
\paragraph{HAP Safe-Subspace Prior}
Let $\mathcal{C}_{\mathrm{safe}}$ denote the set of safe Cartesian positions derived from the HAP traversable subspace.
Given a sparse set of queried Cartesian positions 
$\{\mathbf{x}_q\}_{q=1}^{N_q}$ sampled along $\boldsymbol{\zeta}'(t)$,
we minimise a soft-min distance to $\mathcal{C}_{\mathrm{safe}}$:
\begin{equation}
\mathcal{L}_{\mathrm{hap}}
=
\frac{1}{N_q}
\sum_{q=1}^{N_q}
-\tau \log
\sum_{\mathbf{y} \in \mathrm{kNN}(\mathbf{x}_q)}
\exp
\left(
-\frac{\|\mathbf{x}_q - \mathbf{y}\|}{\tau}
\right),
\label{eq:hap_compact}
\end{equation}
where $N_q$ denotes the number of sampled trajectory query points and 
$\mathrm{kNN}(\mathbf{x}_q)$ returns the $k$ nearest neighbours in 
$\mathcal{C}_{\mathrm{safe}}$.

\paragraph{Final Actor Objective}
The complete actor loss is
\begin{equation}
\begin{aligned}
\mathcal{L}_\pi
&=
\mathbb{E}\!\left[\alpha \log \pi(\mathbf{a}_{t_s}|\mathbf{s}_{t_s})-Q_{\min}(\mathbf{s}_{t_s},\mathbf{a}_{t_s})\right]\\
&\quad+\lambda_{\mathrm{sim}}\mathcal{L}_{\mathrm{sim}}
+\lambda_{\mathrm{hap}}\mathcal{L}_{\mathrm{hap}} .
\end{aligned}
\label{eq:actor_final_compact}
\end{equation}

This formulation jointly promotes return maximisation, kinematic retention, and safety-aware exploration in via-point space.

\subsection{Safety-aware Policy Exploration}

To embed safety-awareness in policy search, we employ dense distance-based reward shaping to discourage unsafe adaptation 
\begin{equation}
\label{eq:safety_reward}
r_{\mathrm{sd}}^{t_s}
= -\, \beta \sum_{i}
\frac{\mathbf{1}\!\left[d(\mathbf{x}_{t_s}^i) < d_{\min}\right]}
{d(\mathbf{x}_{t_s}^i) + \epsilon},
\end{equation}
where $\mathbf{x}_{t_s}^i$ denotes sampled points on the end-effector at time step $t_s$, $d(\cdot)$ is the EDF, 
$d_{\min}$ is the safety margin, 
$\beta>0$ is a scaling coefficient, and $\epsilon>0$ avoids singularities as $d(\cdot)\!\to\!0$. 
This hyperbolic formulation yields a rapidly increasing penalty near contact, providing a stronger safety gradient compared to linear shaping.

Furthermore, we compute EDF gradients at the active via-points $\boldsymbol{\Gamma}_{t_s}^{\mathcal{W}_{t_s}}$, which define locally safe directions. 
Based on this, we define a gradient-alignment reward
\begin{equation}
\label{eq:grad_alignment_reward}
r_{\mathrm{ga}}^{t_s}
=
\frac{1}{L}
\sum_{j=1}^{L}
\cos\!\left(
\Delta \boldsymbol{\Gamma}_{t_s}^{(j)}, \nabla d(\mathbf{v}_{t_s}^{(j)})
\right),
\end{equation}
where $\mathbf{v}_{t_s}^{(j)}$ denotes the Cartesian position of the $j$-th via-point in $\boldsymbol{\Gamma}_{t_s}^{\mathcal{W}_{t_s}}$, and $\Delta \boldsymbol{\Gamma}_{t_s}^{(j)}$ is the corresponding update component. 
A higher value reflects actions aligned with obstacle-avoidance directions. The complete reward at time step $t_s$ is defined as
\begin{equation}
\label{eq:total_reward}
r_{t_s}
=
r_{tc}^{t_s}
+
r_{\mathrm{sd}}^{t_s}
+
\lambda_{ga} \, r_{\mathrm{ga}}^{t_s},
\end{equation}
where $r_{tc}^{t_s}$ denotes the task completion reward, $\lambda_{ga} > 0$ balances the influence of gradient-aligned exploration. This formulation simultaneously encourages task completion, clearance maintenance, and safety-directed exploration during policy learning. Note that, although the distance penalty and gradient-aligned reward encourage collision-free behaviour, they do not provide formal safety guarantees.
\subsection{Training Details}
Algorithm~\ref{algo:dist_gprl} summarises the training procedure of Dist-GPRL. We initialise the EDF $d(\cdot)$ from the workspace model and update it online according to the moving obstacle state during rollouts. At the beginning of each episode, the demonstrated trajectory is parameterised using a GP and conditioned on the current TC to obtain the initial via-point set $\boldsymbol{\Gamma}_{0}$. 

At each time step $t_s$, the active via-point window $\mathcal{W}_{t_s}$ is determined from the phase variable $\phi_{t_s}$. The policy observes the windowed via-points $\boldsymbol{\Gamma}_{t_s}^{\mathcal{W}_{t_s}}$ together with EDF clearances and task vectors, and outputs a corrective action. The correlated update $\Delta \boldsymbol{\Gamma}_{t_s}$ is applied to the active window via GP conditioning, yielding updated via-points that define the adapted trajectory segment. 
This segment is executed in the environment to obtain rewards and the next state. One transition is stored per time step, and the policy is updated using the safety-shaped reward and the regularised actor objective in Eq.~\eqref{eq:actor_final_compact}.
\begin{algorithm}[H]
\caption{Dist-GPRL: Safety-Aware Skill Adaptation}
\label{algo:dist_gprl}
\begin{algorithmic}[1]
\Require Demonstration $\boldsymbol{\zeta}(t)$
\Ensure Adapted trajectory $\hat{\boldsymbol{\zeta}}(t)$
\State Initialise actor $\pi_\theta$, critics $Q_{\phi_1},Q_{\phi_2}$ and replay buffer
\State Initialise Euclidean Distance Field $d(\cdot)$
\For{each episode}
    \State Reset via-points to demonstration set $\boldsymbol{\Gamma}_0$
    \State Condition GP on observed task configuration $\mathcal{TC}$
    \For{each RL time step $t_s$}
        \State Update EDF using current obstacle state
        \State Determine active window $\mathcal{W}_{t_s}$ from phase $\phi_{t_s}$
        \State Propose corrective update $\tilde{\mathbf{a}}_{t_s} \sim \pi_\theta$
        \State Formulate correlated update $\Delta \boldsymbol{\Gamma}_{t_s} = \mathbf{S}_{\mathcal{W}_{t_s}}\tilde{\mathbf{a}}_{t_s}$
        \State Update via-points and condition GP posterior
        \State Execute adapted trajectory segment
        \State Compute task and safety rewards
        \State Store transition and update SAC parameters
    \EndFor
\EndFor
\State Query GP updated by best policy to obtain $\hat{\boldsymbol{\zeta}}(t)$
\State \Return $\hat{\boldsymbol{\zeta}}(t)$
\end{algorithmic}
\end{algorithm}

\subsection{Hyperparameter Selection}
In our implementation, the GP parameterisation consists of 15 via-points in each axis, resulting in 90 via-points across the six axes. We use three overlapping windows of $L=5$ six-DoF via-points with one-point overlap, resulting in a 30-dimensional action per RL step. Smaller windows limit adaptation flexibility, whereas larger windows approach global adaptation and can reduce learning stability; the overlap maintains continuity between adjacent segments. The covariance transformation uses $\lambda=10^{-2}$, where smaller values retain more of the raw action and larger values produce smoother but more conservative updates. We set $\lambda_{\mathrm{hap}}=0.2$ and $\lambda_{\mathrm{sim}}=0.5$; increasing these values strengthens safe-subspace guidance and kinematic retention, respectively, but excessive regularisation can restrict the deviations required for task adaptation. For the reward function, $\lambda_{ga}$ is set as 0.2 to provide more importance to the task completion while avoiding local collisions. All values were selected empirically and kept fixed across experiments.  

\section{Experiments}
\subsection{Experimental Setup}
\begin{figure}
    \centering
    \includegraphics[width=1\linewidth]{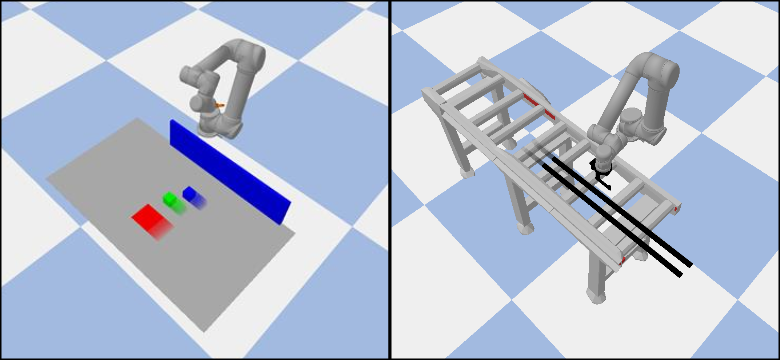}
    \caption{Dynamic cube pushing (left) and dynamic bar removal (right) simulated in PyBullet}
    \label{fig:sim_exps}
\end{figure}

We first validate our approach in simulation using two separate tasks. Both experiments have been conducted in PyBullet \cite{coumans2016pybullet}, with a single expert demonstration for each task. The DBM task is additionally replicated on hardware. In both cases, a Universal Robots UR5e has been used with different custom end-effectors suitable for each task (See Fig.~\ref{fig:sim_exps}). 

\paragraph{Dynamic Cube Pushing (DCP)} The robot equipped with a shovel has to push a moving cube towards a goal region, while avoiding a dynamic obstacle. The cube's initial pose is varied up to 20 cm on the \textit{xy}-plane from the given demonstration. The start pose of the obstacle can also vary up to 10 cm.

\paragraph{Dynamic Bar Manipulation (DBM)} The robot, equipped with a hook end effector, must pick up and remove one of two bars from a moving conveyor belt, while avoiding collisions with the other bar or conveyor rollers. The start poses of both bars also vary up to 20 cm along the \textit{xy}-plane on the conveyor.

We further test the DBM task on hardware where the policy trained on simulation is transferred in one shot. We leverage the IDMP \cite{ali2024interactive} framework to obtain a dynamically updating distance field that is used to query via-point distance values. We use ArUco markers to detect and find the bars' locations in the camera frame. We then transform them to the robot's base frame, utilising an ArUco marker placed near the robot base with a known transform. 

\begin{figure}
    \centering
    \includegraphics[width=1\linewidth]{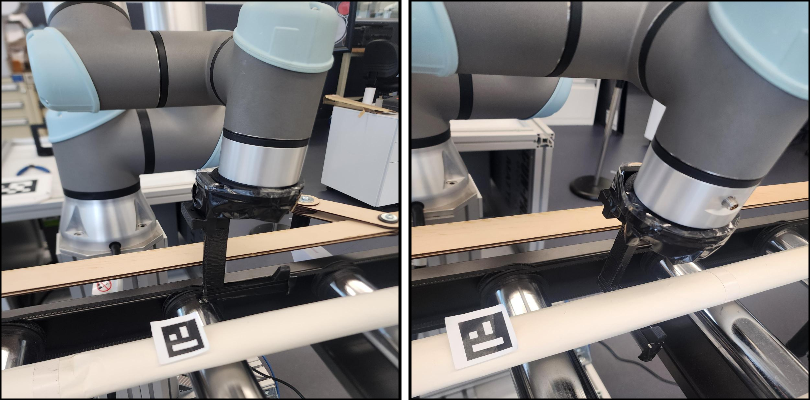}
    \caption{Baseline GPRL fails to avoid collision (left). Our proposed approach avoids collisions and proceeds to pick up the bar successfully (right).}
    \label{fig:hard_exps}
\end{figure}

\subsection{Baselines}
We benchmark our approach against 5 algorithms. 

\textbf{\textit{GPRL}}: The original structured GPRL framework~\cite{haque2026robotskilllearningadaptation}, where SAC directly updates all via-points without windowing, action correlation, or distance-based shaping.

\textbf{\textit{ProMP-RRL}}: Following \cite{carvalho2022adapting}, we fit an object-centric ProMP over 20 demonstrated trajectories, followed by a residual reinforcement learning (RRL) \cite{johannink2018residual} based adaptation over the nominal trajectory offset by the observed object with SAC.

\textbf{\textit{Dist-GPRL-Global}}: Global via-point adaptation with both distance-based reward shaping and HAP prior biasing, but without windowed adaptation or correlation transformation.

\textbf{\textit{GPRL-D}}: Windowed adaptation with correlation transformation and distance-based rewards, but without the HAP safe-subspace prior.

\textbf{\textit{GPRL-H}}: Windowed adaptation with correlation transformation and HAP prior, but without distance-based rewards.

Note that each framework utilises the same SAC configuration as proposed in \cite{haarnoja2018soft}.

\subsection{Simulation Results}
\begin{figure*}
    \centering
    \includegraphics[width=1\linewidth]{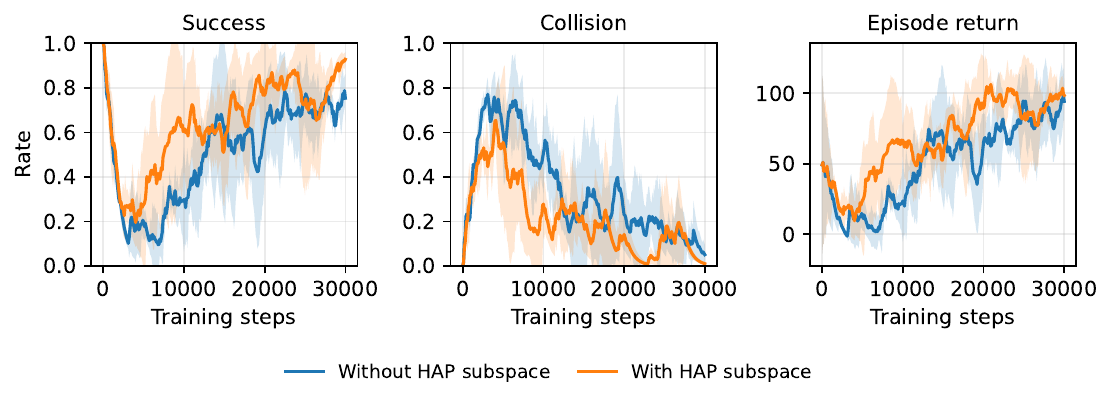}
    \caption{Effect of the HAP subspace on learning across 3 seeds. HAP subspace accelerates convergence, improves final task success, and reduces collision frequency throughout training.}
    \label{fig:hap_effect}
\end{figure*}

\begin{figure*}
    \centering
    \includegraphics[width=1\linewidth]{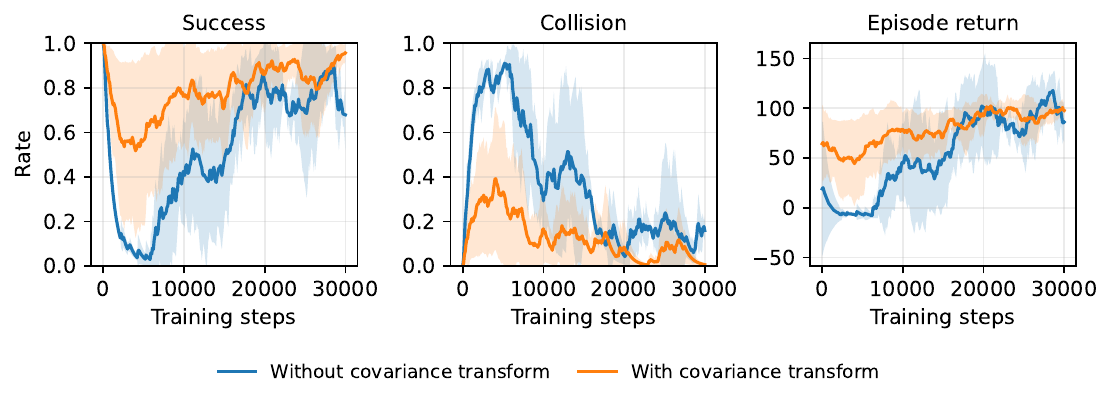}
    \caption{Effect of the covariance-based action transform on learning across 3 seeds. Covariance transformation over the raw actions improves stability with better final task success, reduced collision frequency and higher episodic returns.}
    \label{fig:cov_effect}
\end{figure*}

Our simulated experiment consists of 100 runs with varying TCs, where both the object to be manipulated and the obstacle are moving. The cube pushing task has a wall and a plane as static obstacles, and another moving cube as a dynamic obstacle. Similarly, in bar manipulation, the conveyor rollers and rails act as static obstacles, whereas another moving bar is considered as a dynamic obstacle. The agent must learn to manipulate the objects in question without colliding with any of the static or dynamic obstacles.

Success rate, collision rate, and the 5th-percentile episode-wise minimum clearance are reported in Table~\ref{tab:sim_results}. In both tasks, the baseline GPRL performs poorly with low success rates, the highest collision rates and very low minimum clearance. When we apply our framework and propose updates to all the via-points, and in turn, modify the GP parameterised skill globally, the results showcase improvement over the baseline GPRL. However, the learned policy still leads to collisions and can fail to manipulate the object. 


In contrast, windowed operations perform better in comparison. GPRL-D, operating without the HAP subspace prior, marginally prioritises safety over success. This leads to high minimum clearances, however, with slightly less success. GPRL-H achieves results comparable to GPRL-D; however, on its own, it is not capable enough to achieve state-of-the-art performance. Whereas, our proposed approach achieves the highest success rates, reaching 89\% in DCP and 98\% in DBM. It also attains a 1\% collision rate in DBM and the second-lowest collision rate in DCP (5\%). Furthermore, the method maintains competitive minimum clearance across both tasks, demonstrating its ability to balance task success with safe interaction in cluttered environments.

We further evaluate the impact of the HAP subspace and the covariance-based action transformation on learning performance. All experiments are conducted across three random seeds, and we report the success rate, collision rate, and episode return over training steps. Fig.~\ref{fig:hap_effect} shows that incorporating the HAP subspace consistently yields higher success rates, lower collision frequencies, and higher episode returns compared to the variant without the HAP subspace. A similar trend is observed in Fig.~\ref{fig:cov_effect}, where the proposed covariance-based transformation improves learning stability and sample efficiency, leading to better performance earlier in training compared to the direct action update baseline.


Runtime was measured over the three adaptation windows on an 11th-generation Intel Core i7-1185G7 processor with 16 GB DDR4 RAM. Each window required approximately 4~ms for distance-field queries, 2~ms for local GP conditioning, and 0.5~ms for policy inference, yielding approximately 19.3~ms of computational adaptation time per episode. These results support real-time adaptation feasibility.

\begin{table}[htb]
\centering
\caption{Simulation Results Across Tasks}
\label{tab:sim_results}
\begin{tabular}{C{1.4cm}|C{0.75cm}C{0.75cm}C{0.75cm}|C{0.75cm}C{0.75cm}C{0.75cm}}
\hline
\rule{0pt}{2.2ex}
& \multicolumn{3}{c|}{DCP} & \multicolumn{3}{c}{DBM} \\

Method 
& Success & Collision & MinCl (m)
& Success & Collision & MinCl (m)\\ \hline

GPRL        & 18\% & 68\% & 0.001 & 44\% & 54\% & 0.001 \\ \hline
ProMP-RRL   & 5\% & \textbf{3}\% & \textbf{0.051} & 11\% & 8\% & 0.036 \\ \hline
Dist-GPRL-Global & 64\% & 30\% & 0.004 & 78\% & 11\% & 0.003 \\ \hline
GPRL-D      & 77\% & 9\% & 0.035 & 93\% & 3\% & \textbf{0.04} \\ \hline
GPRL-H      & 79\% & 12\% & 0.01 & 90\% & 7\% & 0.029 \\ \hline
Dist-GPRL (Ours) & \textbf{89}\% & 5\% & 0.017 & \textbf{98}\% & \textbf{1}\% & 0.037 \\ \hline
\end{tabular}
\end{table}

\subsection{Hardware Results}

The hardware experiments on the DBM task across 10 episodes are reported in Table~\ref{tab:hardware_results}. Each trial corresponds to a full execution of the task under randomly initialised bar positions, providing a practical evaluation of robustness under real-world variability. The results exhibit trends similar to those observed in the simulation. Both GPRL and ProMP-RRL achieve poor success rates. GPRL policy fails to avoid collisions and operate within the cluttered space to remove the bar (Fig.~\ref{fig:hard_exps}). ProMP-RRL follows a conservative strategy that avoids congested regions, resulting in very few collisions and relatively high minimum clearance. However, this behaviour leads to minimal interaction with the bar, resulting in zero successful removals. 

Our proposed approach with global adaptation also falls short of the best performance. While it maintains low collision rates, it fails to successfully pick up the bar in half of the trials. Both GPRL-D and GPRL-H achieve moderate success, reflecting their ability to incorporate distance awareness and safe exploration, respectively. However, as also observed in the simulation, combining these components yields the best performance. Our proposed framework achieves the highest success rate while maintaining low collision frequency, demonstrating the benefit of jointly leveraging distance field guidance and structured exploration for reliable real-world manipulation.


\subsection{Discussion}
Experimental results indicate that adapting the full set of via-points leads to high training instability. This is characterised by suboptimal policies that lead to low success rates and high collisions. Despite incorporating the HAP subspace prior and distance-based reward shaping, the agent remains unable to provide actions that result in collision-free paths. A primary driver for this unstable behaviour is the high-dimensional action space since the learnt policy adapts the full set of via-points simultaneously at each update. 
\begin{table}[htb]
\centering
\caption{Hardware Results on the DBM Task}
\label{tab:hardware_results}
\begin{tabular}{C{2.5cm}|C{0.9cm}C{0.9cm}C{0.9cm}}
\hline
\rule{0pt}{2ex}
Method 
& Success & Collision & MinCl (m)\\ \hline

GPRL        & 10\% & 70\% & 0.001 \\ \hline
ProMP-RRL   & 0\%  & 20\% & \textbf{0.03}  \\ \hline
Dist-GPRL-Global & 50\% & 20\% & 0.002 \\ \hline
GPRL-D      & 70\% & 20\% & 0.027 \\ \hline
GPRL-H      & 60\% & 30\% & 0.022 \\ \hline
Dist-GPRL (Ours) & \textbf{90}\% & \textbf{10}\% & 0.027 \\ \hline

\end{tabular}
\end{table}
This introduces a credit assignment problem where the agent cannot discern which specific via-point modifications contributed to a collision or a successful motion, thereby hindering policy convergence. Residual learning leads to opposing results, where the policy finds a suboptimal solution to avoid collision penalties, despite it being positively incentivised towards interacting with the manipulated object. Although ProMP-RRL achieves very low collisions and one of the best minimum clearance values, its success rate is poor.

In contrast, operating locally on nearby via-points of the GP facilitates significantly better credit assignment. Distance-based rewards provide beneficial signals for collision avoidance, whereas the HAP subspace prior enables more efficient exploration and faster convergence, as shown in Fig.~\ref{fig:hap_effect}. Both GPRL-D and GPRL-H outperform global frameworks such as GPRL or Dist-GPRL-global. However, in isolation, these components exhibit specific behaviours; without the HAP prior, distance-based rewards show lower success rates in the cube pushing task, although with a lower number of collisions. Whereas, without the distance field guidance, the HAP subspace alone is insufficient to ensure collision-free behaviour. 

When combined, these mechanisms complement each other. The HAP prior encourages the exploration of safe regions in the action space, facilitating high returns and accelerating convergence (Fig.~\ref{fig:hap_effect}); while distance field rewards provide a precise metric of safety for each action, reducing collisions. 
To further stabilise learning, the covariance-coupled smoothing over the actions ensures that via-points move in unison within each windowed batch. This approach preserves the geometric and kinematic structure of the demonstrated task, leading to better policies during exploration (Fig.~\ref{fig:cov_effect}).

While this work enhances skill adaptation for GP-parameterised trajectories, it also highlights the inherent limitations of GPs, particularly regarding extrapolation. In the DCP task, the adaptation of via-points can accentuate these issues, leading to erratic movements toward the end of the trajectory. While the DCP task is simpler than DBM, its proximity to the plane can often lead to collisions. Future work will investigate strategies to mitigate this issue by stabilising GP extrapolation with additional boundary constraints. Furthermore, the GP conditioning on the new TCs relies on accurate sensor information and does not take into consideration the sensor noise and the uncertainty in the estimated distance field. Future work will take into consideration the uncertainty associated with the sensor measurements and noise. Lastly, since we rely on a learned policy for collision avoidance under varying TCs, our current formulation improves empirical safety but does not provide formal safety guarantees. Future work will look into the prospect of constrained reinforcement learning together with explicit safety mechanisms towards more reliable deployment. We will also explore applications of the proposed framework in real-world industrial applications.

\subsection{Conclusion}

This work presented Dist-GPRL, a safety-aware framework that extends GP-based skill adaptation to cluttered and dynamic environments through local windowed updates, covariance-transformed actions, and complementary HAP-subspace and distance field guidance. We demonstrate that the inclusion of distance-based rewards as well as the HAP subspace prior significantly improves learning and reduces collisions in the presence of moving obstacles. We further showcase how the GP covariance can be used to transform raw RL actions to regularise exploration and attain improved adaptation stability and performance. Across two dynamic manipulation tasks, including real-world robot execution, Dist-GPRL achieved the strongest overall balance between task success, collision avoidance, and kinematic retention among the evaluated methods.

\bibliographystyle{ieeetr}
\bibliography{references}

@inproceedings{nematollahi2022robot,
  title={Robot skill adaptation via soft actor-critic gaussian mixture models},
  author={Nematollahi, Iman and Rosete-Beas, Erick and R{\"o}fer, Adrian and Welschehold, Tim and Valada, Abhinav and Burgard, Wolfram},
  booktitle={2022 International Conference on Robotics and Automation (ICRA)},
  pages={8651--8657},
  year={2022},
  organization={IEEE}
}

@article{ali2024interactive,
  title={Interactive distance field mapping and planning to enable human-robot collaboration},
  author={Ali, Usama and Wu, Lan and M{\"u}ller, Adrian and Sukkar, Fouad and Kaupp, Tobias and Vidal-Calleja, Teresa},
  journal={IEEE Robotics and Automation Letters},
  volume={9},
  number={12},
  pages={10850--10857},
  year={2024},
  publisher={IEEE}
}

@inproceedings{haque2024constrained,
  title={Constrained Bootstrapped Learning for Few-Shot Robot Skill Adaptation},
  author={Haque, AKM Nadimul and Sukkar, Fouad and Tanz, Lukas and Carmichael, Marc G and Vidal-Calleja, Teresa},
  booktitle={2024 IEEE/RSJ International Conference on Intelligent Robots and Systems (IROS)},
  pages={5189--5194},
  year={2024},
  organization={IEEE}
}

@inproceedings{haarnoja2018soft,
  title={Soft actor-critic: Off-policy maximum entropy deep reinforcement learning with a stochastic actor},
  author={Haarnoja, Tuomas and Zhou, Aurick and Abbeel, Pieter and Levine, Sergey},
  booktitle={International conference on machine learning},
  pages={1861--1870},
  year={2018},
  organization={PMLR}
}

@inproceedings{barfoot2014batch,
  title={Batch Continuous-Time Trajectory Estimation as Exactly Sparse Gaussian Process Regression.},
  author={Barfoot, Tim D and Tong, Chi Hay and S{\"a}rkk{\"a}, Simo},
  booktitle={Robotics: Science and Systems},
  volume={10},
  pages={1--10},
  year={2014},
  organization={Citeseer}
}

@article{ijspeert2013dynamical,
  title={Dynamical movement primitives: learning attractor models for motor behaviors},
  author={Ijspeert, Auke Jan and Nakanishi, Jun and Hoffmann, Heiko and Pastor, Peter and Schaal, Stefan},
  journal={Neural computation},
  volume={25},
  number={2},
  pages={328--373},
  year={2013},
  publisher={MIT Press One Rogers Street, Cambridge, MA 02142-1209, USA journals-info~…}
}

@misc{coumans2016pybullet,
  title={Pybullet, a python module for physics simulation for games, robotics and machine learning},
  author={Coumans, Erwin and Bai, Yunfei},
  year={2016}
}

@article{davchev2022residual,
  title={Residual learning from demonstration: Adapting dmps for contact-rich manipulation},
  author={Davchev, Todor and Luck, Kevin Sebastian and Burke, Michael and Meier, Franziska and Schaal, Stefan and Ramamoorthy, Subramanian},
  journal={IEEE Robotics and Automation Letters},
  volume={7},
  number={2},
  pages={4488--4495},
  year={2022},
  publisher={IEEE}
}

@article{paraschos2013probabilistic,
  title={Probabilistic movement primitives},
  author={Paraschos, Alexandros and Daniel, Christian and Peters, Jan R and Neumann, Gerhard},
  journal={Advances in neural information processing systems},
  volume={26},
  year={2013}
}

@article{zhang2025reactive,
  title={A Reactive Framework for Whole-Body Motion Planning of Mobile Manipulators Combining Reinforcement Learning and SDF-Constrained Quadratic Programming},
  author={Zhang, Chenyu and Sun, Shiying and Liu, Kuan and Zhou, Chuanbao and Zhao, Xiaoguang and Tan, Min and Huang, Yanlong},
  journal={arXiv preprint arXiv:2503.23975},
  year={2025}
}

@inproceedings{liu2023safe,
  title={Safe reinforcement learning of dynamic high-dimensional robotic tasks: navigation, manipulation, interaction},
  author={Liu, Puze and Zhang, Kuo and Tateo, Davide and Jauhri, Snehal and Hu, Zhiyuan and Peters, Jan and Chalvatzaki, Georgia},
  booktitle={2023 IEEE International Conference on Robotics and Automation (ICRA)},
  pages={9449--9456},
  year={2023},
  organization={IEEE}
}

@inproceedings{sarkka2011linear,
  title={Linear operators and stochastic partial differential equations in Gaussian process regression},
  author={S{\"a}rkk{\"a}, Simo},
  booktitle={Artificial Neural Networks and Machine Learning--ICANN 2011, 14-17},
  pages={151--158},
  year={2011},
  organization={Springer}
}

@article{sukkar2025multi,
  title={Multi-query Robotic Manipulator Task Sequencing with Gromov-Hausdorff Approximations},
  author={Sukkar, Fouad and Wakulicz, Jennifer and Lee, Ki Myung Brian and Zhi, Weiming and Fitch, Robert},
  journal={IEEE Transactions on Robotics},
  year={2025},
  publisher={IEEE}
}

@article{reynolds2009gaussian,
  title={Gaussian mixture models.},
  author={Reynolds, Douglas A and others},
  journal={Encyclopedia of biometrics},
  volume={741},
  number={659-663},
  pages={3},
  year={2009},
  publisher={Springer City}
}

@inproceedings{liu2022robot,
  title={Robot reinforcement learning on the constraint manifold},
  author={Liu, Puze and Tateo, Davide and Ammar, Haitham Bou and Peters, Jan},
  booktitle={Conference on Robot Learning},
  pages={1357--1366},
  year={2022},
  organization={PMLR}
}

@article{haque2026robotskilllearningadaptation,
  title   = {Towards Robot Skill Learning and Adaptation with Gaussian Processes},
  author  = {Haque, A. K. M. Nadimul and Sukkar, Fouad and Sujipto, Sheila and Le Gentil, Cedric and Carmichael, Marc G. and Vidal-Calleja, Teresa},
  journal = {arXiv preprint arXiv:2603.01480},
  year    = {2026},
  doi     = {10.48550/arXiv.2603.01480}
}

@inproceedings{carvalho2022adapting,
  title={Adapting object-centric probabilistic movement primitives with residual reinforcement learning},
  author={Carvalho, Jo{\~a}o and Koert, Dorothea and Daniv, Marek and Peters, Jan},
  booktitle={2022 IEEE-RAS 21st International Conference on Humanoid Robots (Humanoids)},
  pages={405--412},
  year={2022},
  organization={IEEE}
}

@article{johannink2018residual,
  title={Residual reinforcement learning for robot control},
  author={Johannink, Tobias and Bahl, Shikhar and Nair, Ashvin and Luo, Jianlan and Kumar, Avinash and Loskyll, Matthias and Ojea, Juan Aparicio and Solowjow, Eugen and Levine, Sergey},
  journal={arXiv preprint arXiv:1812.03201},
  year={2018}
}

@article{long2021learning,
  title={Learning barrier functions with memory for robust safe navigation},
  author={Long, Kehan and Qian, Cheng and Cort{\'e}s, Jorge and Atanasov, Nikolay},
  journal={IEEE Robotics and Automation Letters},
  volume={6},
  number={3},
  pages={4931--4938},
  year={2021},
  publisher={IEEE}
}
\end{document}